\documentclass[10pt,twocolumn,letterpaper]{article}

\usepackage[pagenumbers]{anonymous}
\usepackage{times}
\usepackage{epsfig}
\usepackage{graphicx}
\usepackage{amsmath}
\usepackage{amssymb}
\usepackage{pifont}
\usepackage{booktabs}
\usepackage{arydshln}
\usepackage{color}
\usepackage{colortbl}
\usepackage{subcaption}
\usepackage{caption}
\usepackage{multirow}
\usepackage{float}
\usepackage{rotfloat}
\usepackage{capt-of}
\usepackage{longtable}
\usepackage{diagbox}
\usepackage{makecell}
\usepackage{pgfplots}
\usepackage{xspace}
\usepackage{wrapfig}
\usepackage{enumitem}
\usepackage{enumitem}
\usepackage{wrapfig}
\pgfplotsset{compat=1.18}
\usepackage[hypcap=false]{caption} 
\usepackage{inconsolata}

\definecolor{cblue}{rgb}{0.21,0.49,0.74}
\usepackage[pagebackref,breaklinks,colorlinks,allcolors=cblue]{hyperref}
\usepackage[capitalize]{cleveref}
\crefname{section}{Sec.}{Secs.}
\Crefname{section}{Section}{Sections}
\Crefname{table}{Table}{Tables}
\crefname{table}{Tab.}{Tabs.}
\usepackage{url}
\usepackage{afterpage}

\title{STAR: Scale-wise Text-conditioned AutoRegressive image generation}

\author{%
  Xiaoxiao Ma{$^{1,3}$\thanks{Equal contribution; $^{\dagger}$ Corresponding author.}}\,\space\space\space
  Mohan Zhou$^{2,3*}$,\space\space\space
  Tao Liang$^{3}$,\space\space\space
  Yalong Bai$^{3}$,\space\space\space
  Tiejun Zhao$^{2}$,\space\space\space
  Biye Li$^{3}$,\space\space\space
  \\
  Huaian Chen$^{1\dagger}$,\space\space\space
  \vspace{0.15cm}
  Yi Jin$^{1\dagger}$,\space\space\space
  \\
  $^{1}$\normalfont{University of Science and Technology of China}\quad
  $^{2}$\normalfont{Harbin Institute of Technology}\quad
  $^{3}$\normalfont{Du Xiaoman}\quad
  \vspace{0.15cm}\\
    {\tt\small \{xiao\_xiao,anchen\}@mail.ustc.edu.cn, \{mhzhou99,ylbai\}@outlook.com, liangtao@duxiaoman.com} \\
    {\tt\small tjzhao@hit.edu.cn, libiye@gmail.com, jinyi08@ustc.edu.cn}
}

\begin{document}
\twocolumn[{%
    \renewcommand\twocolumn[1][]{#1}%
    \maketitle
    \setlength{\abovecaptionskip}{0.1cm}
    \setlength{\belowcaptionskip}{0.1cm}
    \begin{center}
    \centering
    \vspace{-0.6cm}
    \begin{tabular}{c@{\extracolsep{0.0em}}c} 
	\includegraphics[width=0.95\textwidth]{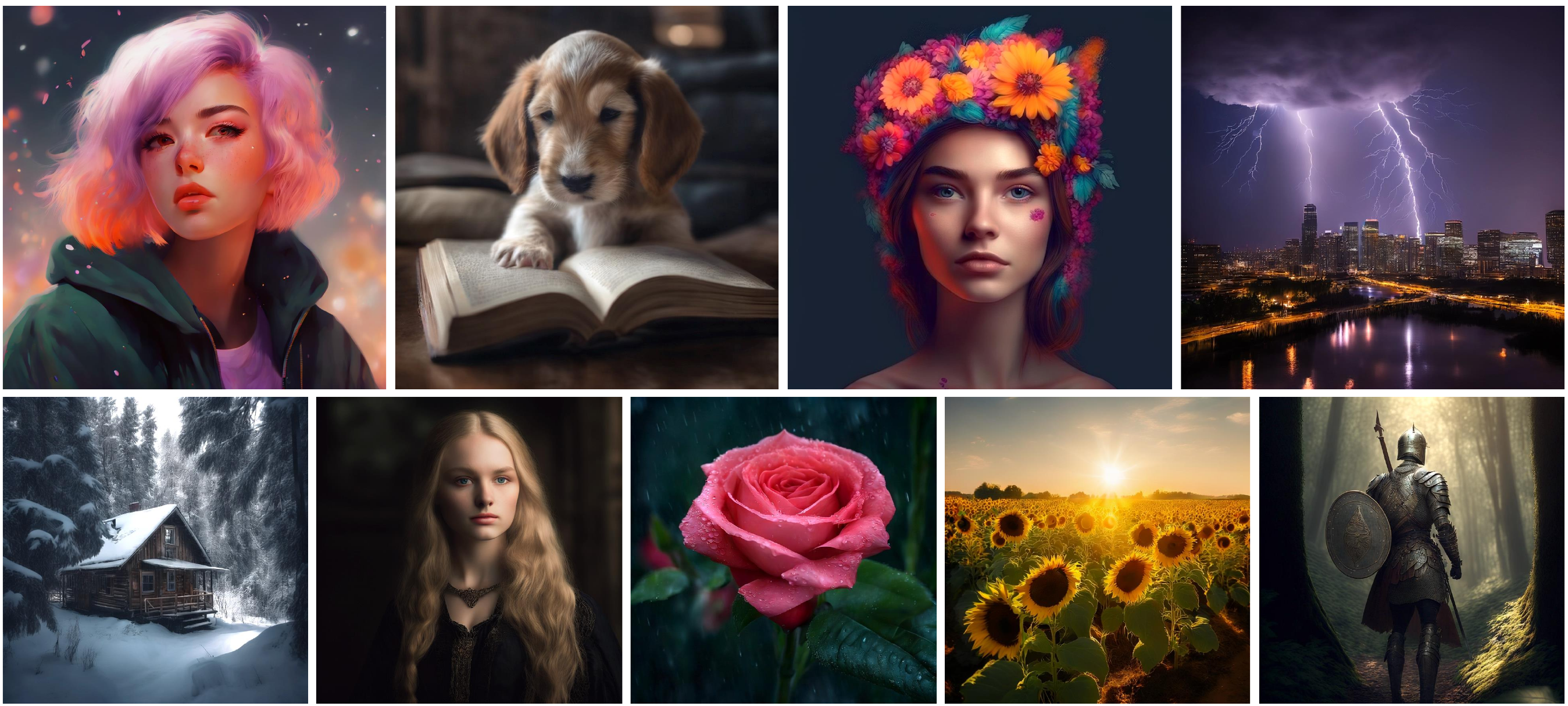}
    \end{tabular}
    \captionof{figure}{STAR directly produces 1024$\times$1024 images with remarkable quality, achieves 3.8$\times$ (measured on A100)  inference speed compared to SDXL. 
    The generated samples demonstrate exceptional detail and fidelity.}
\label{fig:image_cases}
    \end{center}     
}]
\renewcommand{\thefootnote}{\fnsymbol{footnote}}
\footnotetext[1]{~indicates equal contributions.}
\footnotetext[2]{~Corresponding authors.}
\begin{abstract}
\label{abstract}
We introduce STAR, a text-to-image model that employs a scale-wise auto-regressive paradigm. Unlike VAR, which is constrained to class-conditioned synthesis for images up to 256$\times$256, STAR enables text-driven image generation up to 1024$\times$1024 through three key designs.
First, we introduce a pre-trained text encoder to extract and adopt representations for textual constraints, enhancing details and generalizability.
Second, given the inherent structural correlation across different scales, we leverage 2D Rotary Positional Encoding (RoPE) and tweak it into a normalized version, ensuring consistent interpretation of relative positions across token maps and stabilizing the training process.
Third, we observe that simultaneously sampling all tokens within a single scale can disrupt
inter-token relationships,
leading to structural instability, particularly in high-resolution generation. To address this, we propose a novel stable sampling method that incorporates causal relationships into the sampling process, ensuring both rich details and stable structures.
Compared to previous diffusion models and auto-regressive models, STAR surpasses existing benchmarks in fidelity, text-image consistency, and aesthetic quality, requiring just 2.21s for 1024$\times$1024 images on A100. This highlights the potential of auto-regressive methods in high-quality image synthesis, offering new directions for the text-to-image generation. Available at~\url{https://github.com/Davinci-XLab/STAR-T2I/}.
\end{abstract}
\section{Introduction}
\label{introduction}
\begin{figure*}[ht]
\centering
\noindent
\begin{minipage}[t]{1.3\columnwidth}
\vspace{-2mm}
\input{figs/chart_sota}
\end{minipage}%
\hfill
\begin{minipage}[t]{0.68\columnwidth}
\vspace{-3mm}
\setlength{\abovecaptionskip}{0.1cm}
\setlength{\belowcaptionskip}{0.1cm}
\centering
  \centering
   \includegraphics[width=1\linewidth]{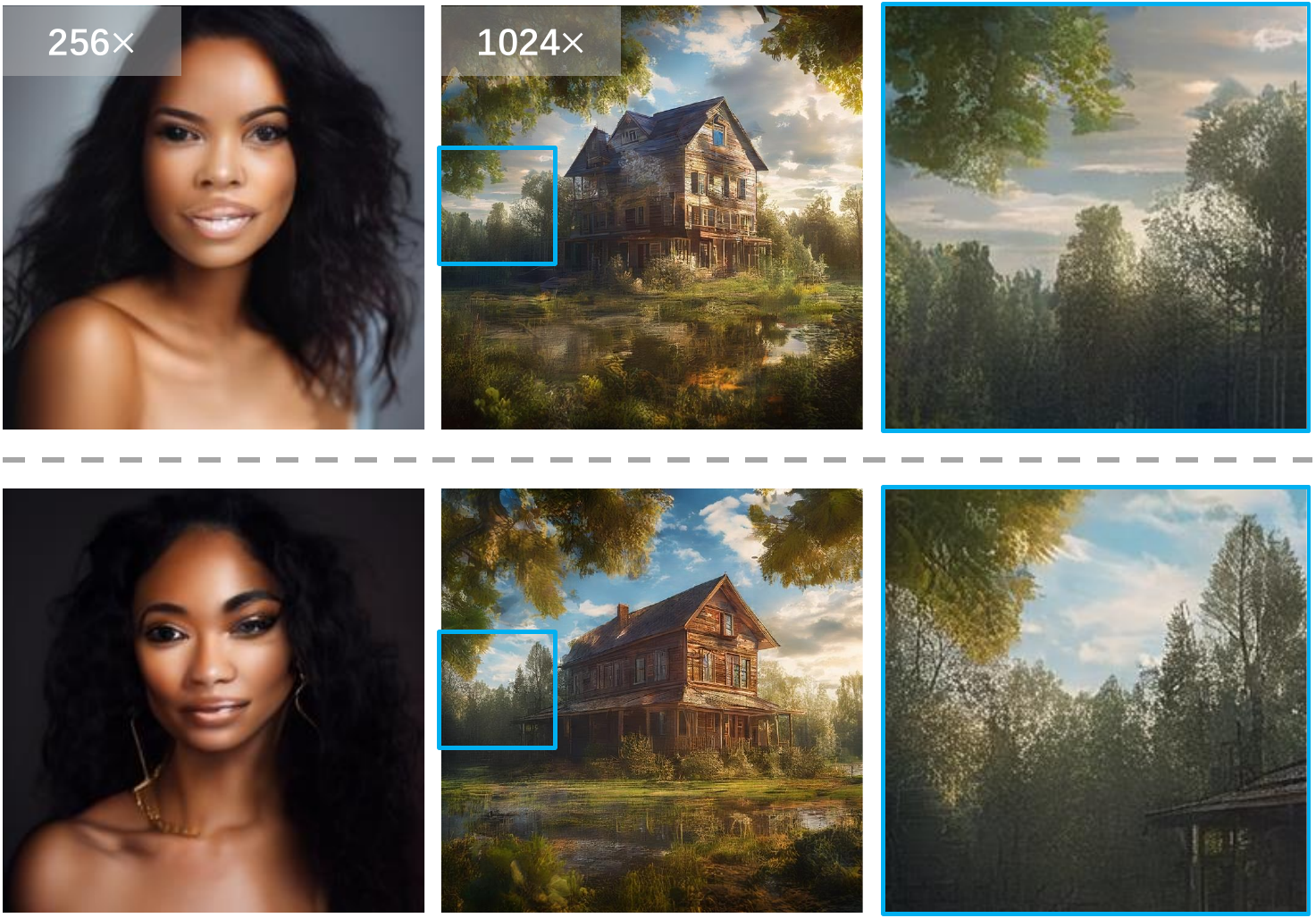}
   \captionof{figure}{Comparison with sampling strategy in~\cite{tian2024var} (top) and our proposed (bottom). Sampling all tokens at once can cause structural instability, which our approach mitigates while preserving rich image details.}
   \vspace{-0.5cm}
   \label{fig_intro_sample}
\end{minipage}
\vspace{-3mm}
\end{figure*}
Visual generation has become a key research area in computer vision community, dominated by diffusion models like Stable Diffusion~\cite{rombach2022stablediffusion,esser2024sdv3}, FLUX~\cite{flux_github}, and~\etc. 
These models can produce high-quality, high-resolution outputs through a progressive denoising process.
However, diffusion models continue to face criticism for their slow denoising speed. Despite efforts to accelerate sampling through distillation schemes~\cite{luo2023latent} and efficient sampler~\cite{lu2022dpm}, these approaches often trade speed gains for image quality.

Inspired by Large Language Models (LLMs)~\cite{touvron2023llama,touvron2023llama2}, Auto-regressive (AR) models~\citep{yu2022parti} have shown effectiveness in visual synthesis~\cite{esser2021vqgan,liu2024lumina_mgpt,he2024mars}.
Typically, AR models adopt discrete tokenizers~\cite{van2017vqvae,esser2021vqgan} to quantize image and employ transformers to predict tokens sequentially. 
This process is highly time-intensive due to the large number of tokens required for high-resolution generation, and may suffer potential degradation, as images are inherently highly-structured and require bidirectional, 2D dependencies.

Recently, Tian~\etal~\cite{tian2024var} introduce VAR, a scale-wise paradigm that shifts the generation process from ``next-token prediction" to ``next-scale prediction", predicting an entire token map at each scale rather than a single token per forward pass. This structured encoding of image content reduces inference costs and preserves high image quality, markedly enhancing scalability and computational efficiency over prior AR and diffusion models.

However, the capability of VAR under textual instructions or complex, high-resolution images remains unverified. 
First, VAR initiates the entire generation process with a special start token, specifically a category embedding for ImageNet~\cite{deng2009imagenet} generation. This approach, however, falls short for producing images with detailed textual guidance. 
Second, VAR learns new position embeddings for each token and each scale and overlooks cross-scale token correlations within the image pyramid, which can complicate training for new resolutions and hinder high-resolution image generation.
Additionally, unlike traditional AR models, VAR simultaneously generates and samples entire scale of tokens. As shown in~\cref{fig_intro_sample}, the token sampling directions within each scale may vary independently, which can introduce instability during generation, especially for complex or intricate scenes, thereby limiting image quality.

To address these challenges, we introduce STAR, an efficient \textbf{S}cale-wise \textbf{T}ext-conditioned \textbf{A}uto-\textbf{R}egressive framework. To unleash the power of the start token and enable natural language guidance, we first integrate a pre-trained text encoder to process textual inputs, yielding 1) a compact representation as the start token to guide the overall image structure and 2) detailed representations as supervised signals, providing nuanced textual guidance for precise text-conditioned generation. To further strengthen the correlation across multiple scales, we replace the absolute position embedding in VAR with a normalized 2D Rotary Positional Encoding (2D-RoPE)~\cite{su2024roformer}, which improves both training efficiency and stability while also boosting model scalability and convergency speed by enabling progressive training from lower resolutions. Moreover, we also introduce a novel \textit{causal-driven stable sampling} strategy, which learns inner-token relationships through a token-level self-supervised training process. This approach provides more accurate guidance during sampling process, resulting in improved stability and coherence in generated images. By better capturing token interactions, we enhance structural integrity, as shown in~\cref{fig_intro_sample}.
As a result, STAR achieves high fidelity and superior efficiency compared to diffusion models (\cref{fig:chart_sota}
), while also generating high-resolution (1024$\times$1024) images with enriched details (\cref{fig:image_cases}).

The main contributions can be summarized as follows:
\begin{enumerate}
    \item We propose STAR, a new auto-regressive paradigm, which empowers the scale-wise paradigm introduced by VAR with textual features and normalized RoPE for high-quality T2I generation at 1024 resolution.
    \item We propose a causal-driven stable sampling strategy that learns inner-token relationships through self-supervised training, enhancing stability and coherence for token-based image generation.
    \item Extensive experiments and qualitative analysis are conducted to demonstrate the superiority of STAR over current methods. STAR achieves remarkable performance in fidelity, text-image consistency, particularly in producing highly detailed images with more efficiency.
\end{enumerate}
\section{Related Works}
\label{related_works}

Visual generation is now dominated by diffusion models \cite{Dalle-2,saharia2022imagen,rombach2022stablediffusion,dhariwal2021diffusion}, surpassing techniques like GANs~\cite{xu2018attngan,tao2020dfgan,kang2023gigagan} and VAEs~\cite{kingma2013auto}, especially in text-to-image generation. Inspired by Large Language Models(LLMs)~\cite{touvron2023llama,touvron2023llama2}, autoregressive approaches for visual generation have increasingly garnered attention, owing to their swift generation speed and inherent scaling capabilities.

\noindent{\textbf{Diffusion models.}}
Latent diffusion model~\cite{rombach2022high, podell2023sdxl} applies U-Net to progressively denoise a Gaussian noise and generate an image. Later, DiT~\cite{peebles2023dit}, PixArt~\cite{chen2023pixart_alpha,chen2024pixart_sigma,chen2024pixart_delta} replace the U-Net with transformers, leveraging the superior scaling capabilities of transformers.
Recently, \cite{liu2024playground, flux_github, esser2024scaling} further scaled up text-to-image diffusion models to billions of parameters in pursuit of higher generation quality.
They have achieved significant progress across various benchmarks, and can produce high-quality, high-resolution outputs through a progressive denoising process.
However, diffusion models continue to receive criticism for slow denoising speed. 
Despite efforts like distillation~\cite{luo2023latent} and efficient samplers~\cite{lu2022dpm} to reduce sampling steps, these techniques often accelerate the process at the cost of image quality.

\noindent{\textbf{Autoregressive (AR) models.}}
For image synthesis, AR models use discrete tokenizers to quantize images~\cite{esser2021vqgan,yu2021vit_vqgan} and transformers to predict tokens sequentially~\cite{gafni2022make_a_scene,yu2022parti,ramesh2021dalle}, instead of denoising an entire latent feature map.
For the complex spatial structure inherent in images, this token-by-token approach does not adhere to the autoregressive assumption, leading to suboptimal results.
VAR~\cite{tian2024var} innovatively shifted next-token prediction to next-scale prediction based on~\cite{lee2022rqvae}, which predicts all tokens within a specific scale aids in maintaining internal consistency within the image, enabling AR models to surpass diffusion models, though currently limited to class-conditioned tasks. 
Expanding on this approach, Li \etal~\cite{li2024controlvar} and Li \etal~\cite{li2024controlar} develop controllable framework; Li~\etal~\cite{li2024imagefolder} further enhance the tokenizer with semantic information.

Unlike diffusion models, sampling strategies play a crucial role in AR approaches.
In large language models (LLMs)\cite{touvron2023llama,touvron2023llama2}, the transformer generates logits for each token, followed by sampling strategies like greedy search, beam search, or top-k/top-p~\cite{holtzman2019top_p}.
For visual generation, LlamaGen~\cite{sun2024llamagen} and Lumina-mGPT~\cite{liu2024lumina_mgpt} emphasize using a significantly higher top-k compared to LLMs to avoid overly smooth outputs and replicated details while increasing randomness.
MaskGIT~\cite{chang2022maskgit} uses mask scheduler to predict new tokens for the final image, and Jose \etal~\cite{lezama2022maskgit_critic} introduce token critic to improve quality by distinguishing VAR~\cite{tian2024var} uses a residual quantizer for parallel token generation with top-k and top-p sampling. However, sampling within the same scale may weaken inter-token correlations, causing instability, especially in high-resolution generation.

\section{Preliminaries}
\label{method_preliminary}
\noindent\textbf{Next-token prediction} is central to traditional auto-regressive models, where images are tokenized into a sequence $(x_1,x_2,...,x_T)$ using VQ-VAE~\cite{van2017vqvae}. The model predicts each token based on the previous ones $p(x_t\mid x_1, x_2, \cdots, x_{t-1})$ and reconstructs the image through VQ-VAE decoders. However, this approach ignores spatial relationships between tokens, predicting them sequentially without explicitly modeling spatial structure.

\begin{figure*}[t]
\setlength{\abovecaptionskip}{0.1cm}
\setlength{\belowcaptionskip}{0.1cm}
\begin{center}
\includegraphics[width=0.8\textwidth]{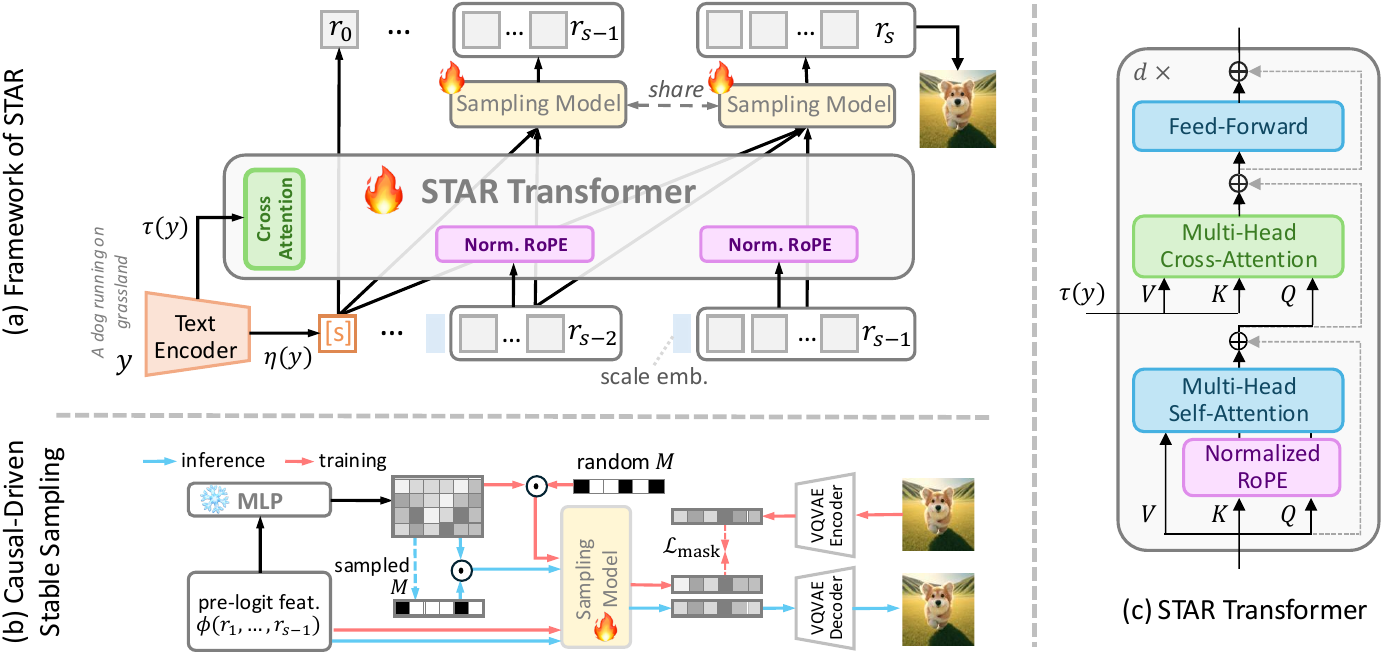}
\end{center}
\caption{
Illustration of STAR.
(a) Given a text prompt, STAR generates images with a compact global representation from a pre-trained text encoder and trains a transformer with Normalized RoPE to gradually predict token maps of higher resolution at a scale-wise manner. At each scale, detailed intermediate representations are infused through cross-attention to boost semantic understanding, resulting in diverse images. (b) To reduce instability from inconsistent sampling directions across scales in high-resolution generation, we have trained a Causal-Driven token sampler and adopted a progressive sampling during inference to synthesize structurally stable and detail-rich images.
}
\label{fig_framework}
\vspace{-3mm}
\end{figure*}

\noindent\textbf{Next-scale prediction.} 
Tian \etal~\cite{tian2024var} argue that "next token prediction" used in language processing is inadequate and inefficient for images due to the highly structured and bi-directional dependency inherent in image structure. They propose a scale-wise auto-regressive paradigm with a multi-scale residual VQ-VAE, beginning with a $1\times 1$ token map $r_1$ and progressively predicts larger-scale maps $(r_1, r_2, \dots, r_S)$.
The process can be formulated as a multiplication of $S$ conditional probabilities:
\begin{equation}
\label{eq:var_cond}
    p(r_1,r_2,...,r_S)=\prod \limits_{s=1}^S p(r_s \mid r_1, r_2, \dots, r_{s-1}),
\end{equation}
where the $s$-th scale's token map $r_s \in [V]^{h_s \times w_s}$ is generated based on previous ones $\{r_1,r_2,...r_{s-1}\}$. Here, $V$ represents the VQ-VAE codebook, while $h_s$ and $w_s$ are the height and width of the token map.

For the $s$-th scale, the model employs cross-entropy loss to minimize the negative log-likelihood of generated tokens. 
During generation, tokens are simultaneously produced according to the conditioned probability distribution $p(r_s\!\mid\!r_{<s})$. The model can use either top-$k$ sampling, selecting the top $k$ highest-probability candidates, or top-$\texttt{p}$ sampling, choosing the smallest set of candidates whose cumulative probability exceeds $\texttt{p}$. In both cases, the selection is weighted by the candidates' probabilities.

VAR achieves SOTA performance in category-based image synthesis, demonstrating improved scalability and efficiency. This breakthrough underscores the potential of auto-regressive models in high-quality image generation.
Nevertheless, challenges remain, including limitations in category-based generation, neglect of cross-scale token correlations, and ongoing concerns about sampling stability. These factors continue to constrain VAR's performance at higher resolutions and in more complex scenarios.

\section{Method}
\label{method}
\subsection{Textual Guidance}
\label{sec_autoregressive_transformer}
Textual guidance plays a crucial role in controllable image generation. VAR uses category embedding as the start token for class-conditioned generation. In this work, we introduce a hybrid textual guidance approach to extend VAR's capabilities, ensuring more diverse scenarios and better consistency between text descriptions and visual outputs.

Specifically, given a natural language input $y$ that encapsulates user preferences for image generation, our STAR employs a pre-trained text encoder $\tau$ to derive a detailed intermediate representation $\tau(y) \in \mathbb{R}^{M \times d_\tau}$, where $M$ denotes the sequence length and $d_\tau$ represents the encoding dimensionality, alongside an aggregation function $\eta(\cdot)$ that generates a compact global representation $\eta(y)$. While $\tau(y)$ facilitates fine-grained conditional signals throughout the generative process, $\eta(y)$ encapsulates high-level structural intentions from the natural language input $y$. This hybrid representation methodology enables the STAR framework to generate high-fidelity text-conditioned images that correspond precisely to user-specified preferences.

As illustrated in~\cref{fig_framework} (a), we utilize the CLIP pooling feature as $\eta(y)$, which serves as an initial token to guide the scale-wise generation process. Simultaneously, we extract $\tau(y)$ through CLIP's text embedding to capture fine-grained textual details. Inspired by the demonstrated effectiveness of cross-attention mechanisms in diffusion models~\cite{rombach2022stablediffusion,chen2023pixart_alpha}, we introduce extra cross-attention layers for $\tau(y)$ between self-attention and feed-forward layers and remove all AdaLN in~\cite{tian2024var} for finer-grained textual guidance at each scale. This architecture enables comprehensive integration of both global structural guidance and fine-grained textual conditions throughout the generation pipeline.

\subsection{Positional Encoding}
We further extend our investigation to optimize the model architecture for high-resolution image generation, focusing on training efficiency and preservation of semantic fidelity at increased spatial dimensions. While VAR utilizes independent absolute positional encoding per scale, neglecting cross-scale token correlations and requiring redundant semantic ``re-training'' at each scale, STAR implements a positional encoding scheme that explicitly captures cross-scale correlations, facilitating efficient multi-resolution semantic learning and enhanced scalability.

Our positional encoding comprises within-scale and cross-scale embeddings. Within-scale embeddings normalize token maps to uniform spatial dimensions, preserving semantic consistency across scales. Cross-scale embeddings use absolute positional encoding to denote scale membership, complementing detailed representation at each scale. This dual encoding paradigm effectively models cross-scale semantic relationships while maintaining scale-specific spatial coherence.

Specifically, we implement a normalized 2D Rotary Position Encoding (RoPE) for the within-scale positional embeddings. For a given scale $s$, we define a 2D grid of dimensions $h_s \times w_s$. At coordinates $(i, j)$, where $i \in \{1,2,\ldots,h_s\}$ and $j \in \{1,2,\ldots,w_s\}$, the positional encoding $\text{PE}(i, j)$ is formulated as:
\begin{equation}
\text{PE}(i, j) = \text{RoPE}_x\left(\frac{i}{h_s}\cdot H\right) \oplus \text{RoPE}_y\left(\frac{j}{w_s}\cdot W\right),
\end{equation}
where $H$ and $W$ represent normalized grid dimensions, with constraints $H \geq h_S$ and $W \geq w_S$. For implementations targeting a maximum resolution of 1024 pixels, we set $H = W = 1024/16$. Here, $\oplus$ represents concatenation along the channel dimension, and $\text{RoPE}_x(\cdot)$ and $\text{RoPE}_y(\cdot)$ denote the rotary embeddings for horizontal and vertical dimensions, respectively. 

\begin{figure}[t]
\centering
\setlength{\abovecaptionskip}{0.1cm}
\setlength{\belowcaptionskip}{0.1cm}
\includegraphics[width=0.45\textwidth]{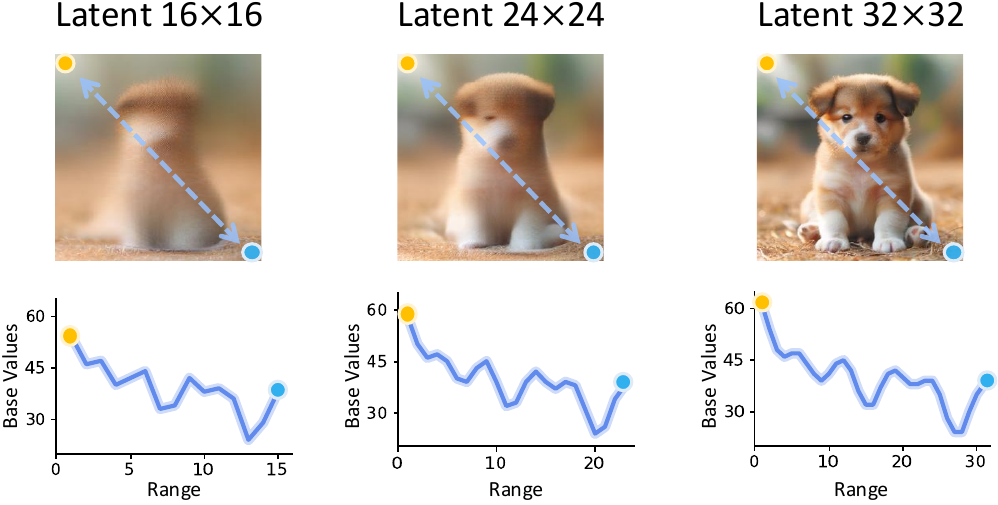}
\caption{Although token maps at different scales have different sizes, their relative positions convey the same meaning (see above). With normalized relative position encoding, models comprehend all scales from a unified perspective (see below).}
\label{fig_rope_scheme}
\vspace{-3mm}
\end{figure}

As demonstrated in~\cref{fig_rope_scheme}, the incorporation of within-scale positional embeddings enables the model to develop a unified comprehension across multiple scales of representation, this is crucial for establishing robust cross-scale dependencies and allowing the model to efficiently scale from lower to higher resolution configurations.

\subsection{Causal-Driven Stable Sampling}
\label{sec_sampling_strategies}
While our model demonstrates the capability to synthesize high-fidelity images through the integration of hybrid textual guidance and dual position encoding mechanisms, an important theoretical question emerges regarding the optimality of token selection strategies: given the logit distribution $p(r_s)$ for image token generation, conventional sampling methods, such as top-$\texttt{p}$ and top-$k$, is insufficient due to the fundamental distinctions between natural language sequences and hierarchical image representations.

\begin{figure}[!t]
\begin{center}
\includegraphics[width=\linewidth]{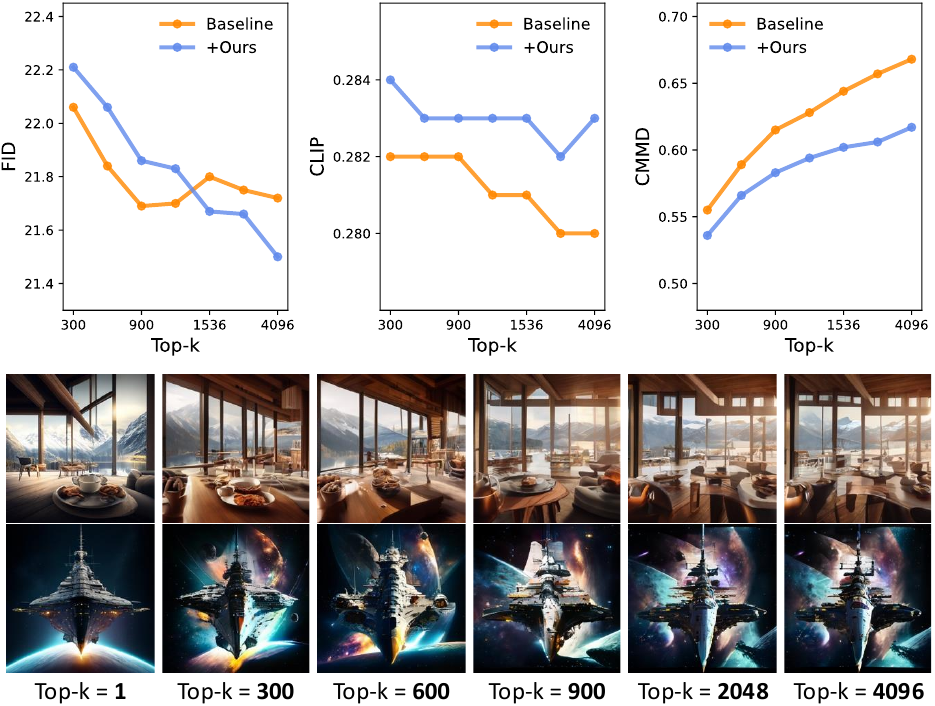}
\end{center}
\vspace{-3mm}
\caption{
Illustration of the impact of varying top-k values on text-image consistency, detail enrichness, and image realism. A lower top-k can lead to better text-image consistency and clearer details, i.e. better CLIP and CMMD~\cite{jayasumana2024cmmd}; A larger top-k can obtain more realistic images, while may generate chaotic details. CMMD is a robust image quality metric using richer CLIP embeddings and Gaussian RBF to capture and assess distortion levels in images.
}
\label{fig_illustrate_sample}
\vspace{-3mm}
\end{figure}

Specially, while VAR enables parallel generation of tokens within each scale, independent sampling across scales can lead to inconsistent generation trajectories, with errors in previous scales propagating through the hierarchical dependency structure, leading to visual distortions in complex scenes. As shown in~\cref{fig_illustrate_sample}, modulating sampling stochasticity via top-$k$ values yields an interesting trade-off between structural coherence and richness. Our analysis reveals that increasing the $k$ values leads to several degradations, including a diminished text-image alignment, as measured by CLIP-score, and a marked deterioration in structural integrity, quantified by elevated CMMD. This empirical evidence suggests that the sampling strategies should be polished for scale-wise auto-regressive generation.

\begin{table*}[t]
\centering
\setlength{\abovecaptionskip}{0.1cm}
\setlength{\belowcaptionskip}{0.1cm}
\caption{Performance comparison under MJHQ-30K shows that our method achieves comparable performance regarding current SoTA diffusion models, and surpasses recent AR models with a generation time of just 2.21 seconds per image (all experiments are evaluated in FP16 precision on A100.). ``CLIP" denotes CLIP-Score. \textbf{Bold} values denote best performance, and \underline{underlined} values are the second-best.}
\label{tab_fid}
\setlength{\tabcolsep}{3mm}{
\resizebox{1.6\columnwidth}{!}{%
\small
\begin{tabular}{lccccccc}
\toprule
\multirow{2}{*}{Methods} & \multirow{2}{*}{Type} & \multirow{2}{*}{Reso} & \multicolumn{2}{c}{MJHQ-30k~\cite{li2024playground}} & \multirow{2}{*}{GenEval~\cite{ghosh2024geneval}$\uparrow$} & \multirow{2}{*}{\makecell{Infer. \\ Time {[}s{]}}} \\ \cmidrule(lr){4-5}
 &  &  & FID$\downarrow$ & CLIP$\uparrow$ &  &  \\
\midrule
SD v2.1~\cite{rombach2022stablediffusion} & Diff. & 768 & 13.84 & 0.278 & 0.52 & 7.94\\
SD XL~\cite{podell2023sdxl} & Diff & 1024 & 6.54 & 0.287 & 0.55 & 8.48\\
PixArt-$\alpha$~\cite{chen2023pixart_alpha} & Diff & 1024 & \underline{6.10} & 0.286 & 0.48 & \underline{6.43} \\
Playground v2.5~\cite{li2024playground} & Diff & 1024 & 6.49 & \textbf{0.294} & \underline{0.56} & 8.56\\
FLUX.1-dev~\cite{flux_github} & Diff & 1024 & 9.90 & 0.281 & \textbf{0.68} & 25.3\\
LlamaGen~\cite{sun2024llamagen} & AR & 512 & 25.61 & 0.230 & 0.35 & 22.4\\
Meissonic~\cite{bai2024meissonic} & AR & 1024 & 21.3 & 0.279 & 0.52 & 17.8\\
\rowcolor[HTML]{EFEFEF} 
STAR & AR & 1024 & \textbf{5.25} & \underline{0.291} & 0.55 & \textbf{2.21} \\
\bottomrule
\end{tabular}
}}
\end{table*}

Drawing inspiration from mask-based approaches~\cite{chang2022maskgit}, as shown in~\cref{fig_framework}(b), we introduce causality into the sampling process to mitigate this issue. This is achieved by implementing a shallow network, self-supervisedly trained to reconstruct randomly masked tokens. By concatenating features from the STAR transformer's final layer, we optimize $\mathcal{L}_{mask}^{s}$ through a mask-prediction scheme:
\begin{equation}
\begin{aligned}
    \mathcal{L}_{mask}^{s}\!=\!\sum_{i,j,m_{i,j}=1} \log p\left(\tilde{M}\odot r_s^{i,j} \mid M \odot r_s, \phi(r_1,\!\dots\!,r_{s-1})\right),
\end{aligned}
\end{equation}
where $\phi(\cdot)$ denotes the pre-logit features from the STAR transformer backbone, $M=[m_{i,j}], m_{i,j}\in{0,1}$ represents the random mask, $\tilde{M}$ represents element-wise negation. The masking operation $\odot$ replaces tokens with \verb|[MASK]| when $m_i=0$. During inference, this sampler transforms the sampling process into a multi-step procedure, establishing causal dependencies among tokens within the current scale. As illustrated in~\cref{fig_illustrate_sample}, our proposed masking mechanism achieves superior FID scores while maintaining text-image alignment and avoids structural degradation in the increased sampling stochasticity.
By leveraging the confidence of token predictions, part of the AR transformer's output can be used as known information, providing more accurate guidance to the sampler and further improving generation stability.

Furthermore, we explore how to make the sampling process in high-resolution image synthesis more efficient. In the hierarchical structure of the feature map generation procedure, initial scales determine structural composition, while subsequent scales govern fine-grained details. By constraining sampling exclusively to later scales, we substantially reduce computational complexity during inference. Besides, our empirical investigation demonstrates that increased sampling iterations enhance stability, with larger scales necessitating additional sampling steps. Thus, we introduce scale-dependent sampling iterations, yielding improved stability while preserving intricate details.

\subsection{Efficient Optimization Strategy}
\label{sec_sampling_strategies_opt}
Training large-scale generative models for high-resolution image synthesis presents significant computational challenges in computational complexity. We address the challenges through two key innovations in our training methodology: an efficient training procedure that leverages local attention patterns to reduce token count in attention layers and a progressive fine-tuning strategy that enables incremental learning of multi-scale semantics.

\noindent{\textbf{Efficient training.}}
During training, the transformer processes concatenated token maps across all scales, yielding an attention map of size $(\sum_{s=1}^{S} h_s \times w_s)^2$, which is computationally intensive. Leveraging our empirical observation that tokens in later scales exhibit strong local dependencies, we introduce a novel training strategy: randomly cropping and training exclusively on tokens within a local window, thereby reducing the effective token count. This optimization achieves a 1.5$\times$ acceleration in training speed while enabling 2-3$\times$ larger batch sizes. 

\noindent{\textbf{Progressive fine-tuning.}}
To address the challenges posed by complex structures in high-resolution images, we utilize a progressive training strategy, initially training on low-resolution images to efficiently learn fundamental image structures and global compositional patterns, followed by fine-tuning on higher resolutions to acquire the capacity for synthesizing fine-grained details and local textures. This learning paradigm facilitates more efficient model convergence while ensuring the robust capture of both macro structural relationships and micro visual details. As shown in~\cref{fig_ab_rope}, thanks to our dual position encoding design, high-resolution image generation capabilities emerge after merely a few thousand fine-tuning steps, with extended training for further quality improvements.

\begin{figure*}[t]
\setlength{\abovecaptionskip}{0.1cm}
\setlength{\belowcaptionskip}{0.1cm}
\begin{center}
\includegraphics[width=1.0\textwidth]{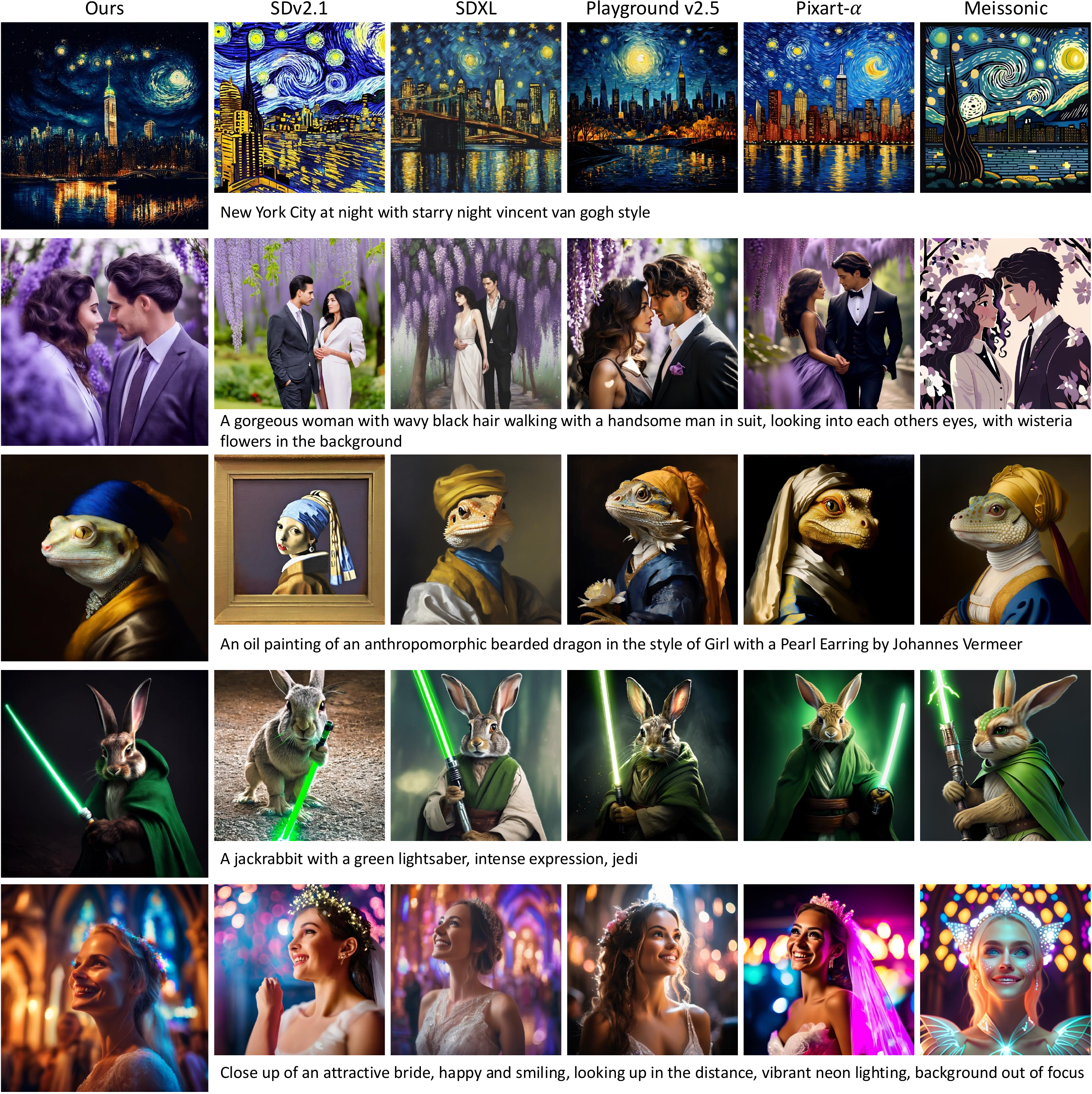}
\end{center}
\caption{
Qualitative Comparison between STAR and other models.
STAR demonstrates its capability to generate high-quality 1024px images in approximately 2.21 seconds (See fig.~\ref{fig:comp_sota}), achieving visual results comparable to SoTA diffusion models~\cite{li2024playground,podell2023sdxl} and outperforming recent autoregressive models~\cite{bai2024meissonic} in both speed and image fidelity.
}
\label{fig:comp_sota}
\vspace{-3mm}
\end{figure*}

\section{Experiments}
\label{experiment}
\subsection{Implementation Details}
\noindent\textbf{Model Configurations.}
Following~\cite{tian2024var}, we construct models with depths of 16 and 30, containing 270M and 1.7B parameters, respectively. 
We observe that features at different scales exhibit discrepancies, which could lead to potential training difficulties. We add QK-norm in each transformer block and additional LayerNorm before the final decoder head.
Currently we use CLIP as text encoder same as~\cite{rombach2022stablediffusion}, note that more powerful language models~\cite{2020t5} could further improve performance.

We thoroughly compare STAR with leading diffusion methods, including Stable Diffusion v2.1~\citep{rombach2022stablediffusion} (``SD v2.1), SDXL~\citep{podell2023sdxl}, Playground v2.5~\citep{li2024playground} and PixArt-$\alpha$~\citep{chen2023pixart_alpha}; along with recent AR models including LlamaGen~\cite{sun2024llamagen} and Meissonic~\cite{bai2024meissonic}.
The models' performance is systematically evaluated using FID for fidelity, CLIP-Score for image-text alignment, and GenEval~\cite{ghosh2024geneval} for assessing generation capabilities in complex, multi-object scenarios.

\noindent\textbf{Training details.}
The training process consists of three stages. We begin with a batch size of 512 with a learning rate of 1e-4 to train 256$\times$256 images, then reduce the batch size to 128 for 512$\times$512 images and to 64 for 1024$\times$1024 images. As shown in~\ref{sec_autoregressive_transformer}, this progressive approach ensures effective training across multiple resolutions.
All models are optimized using AdamW~\cite{loshchilov2017adamw} with weight decay of 0.05 and betas of (0.9,0.95). For the smaller model with depth=16, a batch size of 512 is maintained to balance efficiency and memory usage.

\noindent\textbf{Datasets.}
The training dataset comprises approximately 20M image-text pairs from LAION, supplemented with 10M internal images. The internal images have been re-captioned by~\cite{chen2023sharegpt4v} to improve alignment between visual and textual data. Both the LAION and internal image sets are used to train the 256$\times$256 model, ensuring a diverse and comprehensive dataset for this resolution. However, for 512$\times$512 and 1024$\times$1024 models, only the internal data is utilized as it provides higher-quality and more consistency in annotations needed of high-resolution image generation.

\begin{figure}[t]
\setlength{\abovecaptionskip}{0.1cm}
\setlength{\belowcaptionskip}{0.1cm}
\begin{center}
\includegraphics[width=0.9\linewidth]{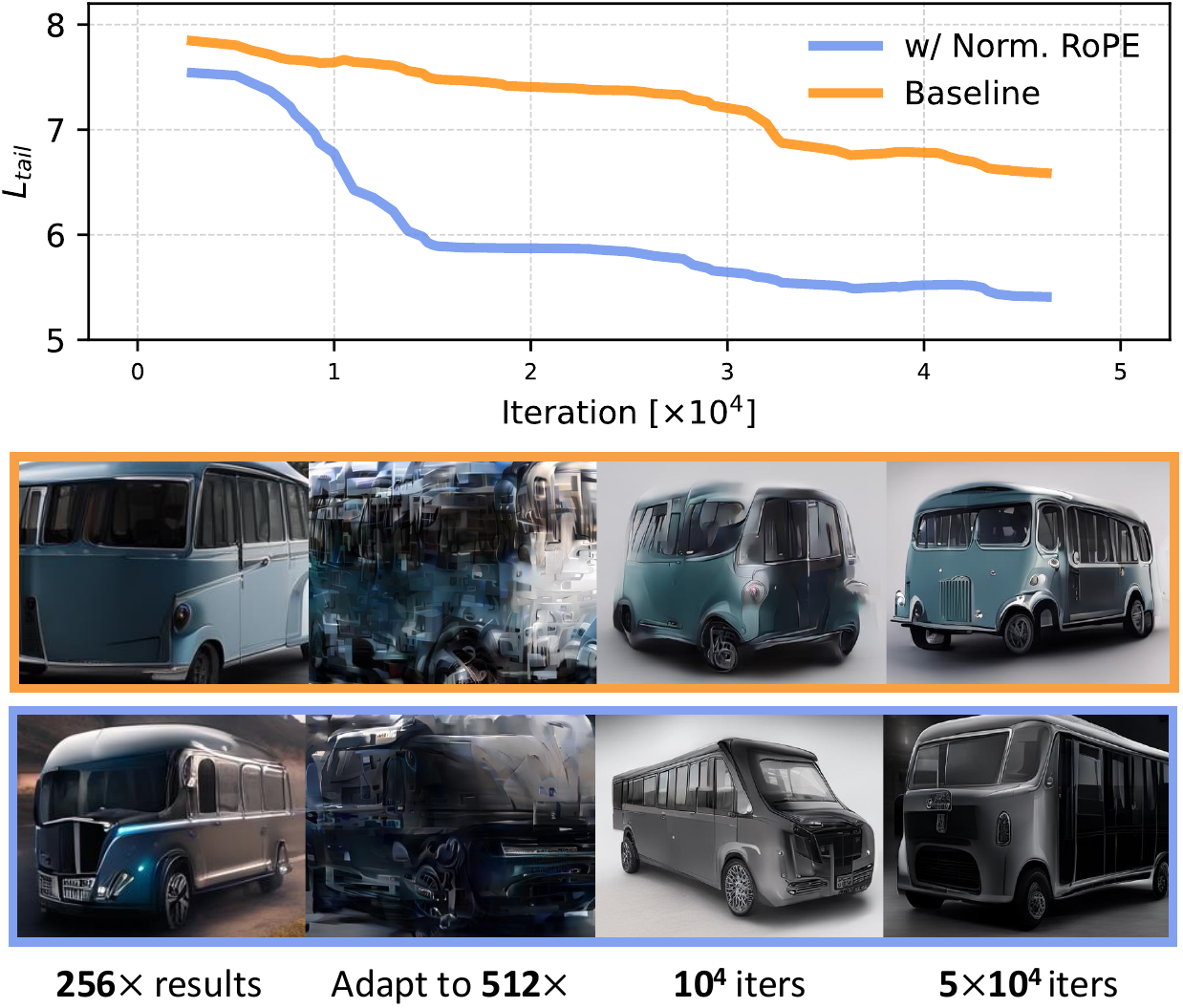}
\end{center}
\caption{
With Normalized RoPE, generative model can adapt to new resolutions with lower cost. To illustrate this, we finetune a 256$\times$ model for 512$\times$ resolution, and report averaged loss of the last scale (upper) along with generated images at different stages (lower). Note that median filtering is applied for visualization.
}
\label{fig_ab_rope}
\vspace{-3mm}
\end{figure}

\subsection{Performances}

We quantitatively assess the fidelity and text-image alignment of STAR’s generated images by measuring both FID and CLIP-score. Due to the stylistic differences between COCO and our generated images, we report results based on MJHQ-30k~\citep{li2024playground}. As shown in Table~\ref{tab_fid}, STAR outperforms in terms of FID, achieving a leading score of 5.25, and demonstrates a competitive CLIP-score of 0.291, highlighting its impressive alignment with textual descriptions and its ability to produce high-quality images.

Visual comparisons of different methods in Fig.~\ref{fig:comp_sota} show that STAR achieves remarkable visual quality comparable to SOTA diffusion models, and even prevails in rendering detailed textures such as fur and fabric. Notably, STAR only requires approximately 2.21 seconds to generate a 1024$\times$1024 image, which is \textgreater3 times faster than existing diffusion and AR models.
This speed advantage, coupled with its visual fidelity, positions STAR as a highly efficient and capable model for image synthesis. The supplementary materials provide additional images and examples.

In addition to visual quality, evaluations on the GenEval benchmark~\cite{ghosh2024geneval} reveal that STAR performs competitively when generating multiple objects in an image.
Notably, STAR outperforms other AR models, such as LlamaGen and Meissonic by 0.2 and 0.03 respectively, showcasing its potential in autoregressive text-to-image generation. While there is still a significant gap compared to FLUX.1-dev, this also highlights ample opportunities for further exploration in text understanding and model scaling for STAR.

\subsection{Analysis \& Ablations}
\noindent\textbf{Transformer -- Positional encodings.}
\label{sec_ab_rope}
We replace RoPE with absolute PE following setting in~\cite{tian2024var} under $depth$=16 setup to illustrate efficiency of our Normalized RoPE. 
Thanks to the unified comprehension across multiple scale representations, as illustrated in Figure~\ref{fig_ab_rope}, the architecture incorporating Normalized RoPE exhibits relatively faster convergence and achieves approximately 10\% higher accuracy throughout the training process compared to the architecture "without Norm.RoPE". Furthermore, the normalized encoding enables more efficient fine-tuning for high-resolution generation based on low-resolution models.

\noindent\textbf{Transformer -- Parameters \& Resolution.}
As shown in Table~\ref{tab_ablation}, increasing $depth$ from 16 to 30, and expanding resolution from 256 to 1024 improves both generation fidelity and text-image alignment. This demonstrates the potential advantage of auto-regressive models in benefiting from large parameters and higher resolutions.

\begin{figure}[t]
\begin{center}
\includegraphics[width=0.9\linewidth]{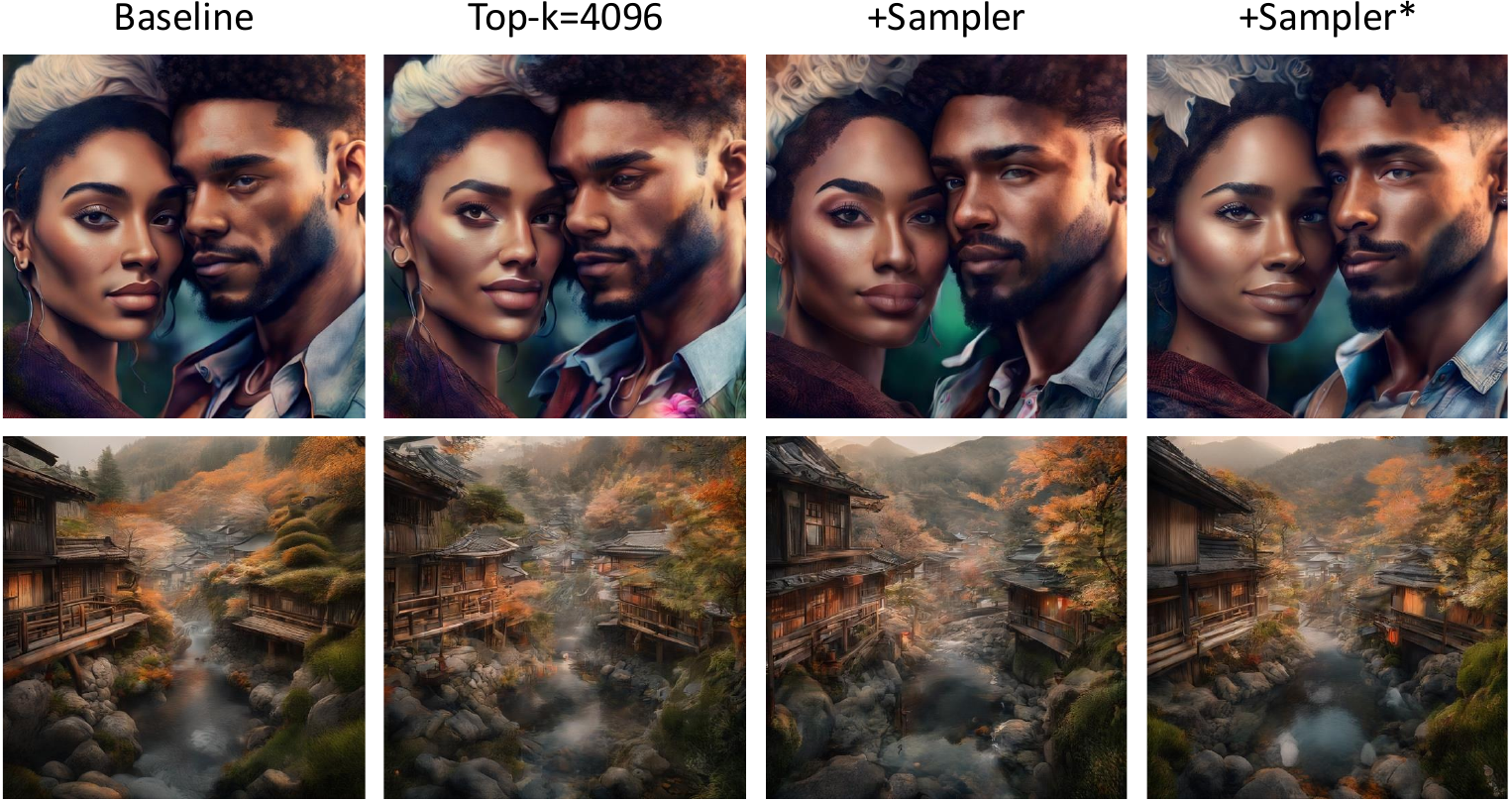}
\end{center}
\vspace{-3mm}
\caption{
Visual comparison of sampling strategies.
``Baseline" uses top-k sampling and k=600. Boosting k to 4096 enhances detail enrichness but introduces structural inconsistencies (\eg, ears, collar in upper and buildings in lower). Establishing causal token relationships as described in Sec.~\ref{sec_sampling_strategies} stabilizes structure while preserving detail (``+Sampler"). Incorporating confidence-based strategies further improves results ("+Sampler*").
}
\label{fig_result_sample}
\vspace{-3mm}
\end{figure}
\begin{table}[tbp]
\setlength{\abovecaptionskip}{0.1cm}
\setlength{\belowcaptionskip}{0.1cm}
\centering
\setlength{\tabcolsep}{2.5mm}{}
\caption{
Results under different sizes (from 256$\times$ to 1024$\times$) and different parameter setups ($depth$=16 and $depth$=30).
}
\label{tab_ablation}
\setlength{\tabcolsep}{2mm}{
\resizebox{1\columnwidth}{!}{
\begin{tabular}{cccccc}
\toprule
Depth & \#Reso & \#Param & CLIP-Score$\uparrow$ & FID$\downarrow$ & Infer. Time \\ \midrule
16   & 256  & 274M  & 0.274  & 6.58 & 1.25 \\
30   & 256  & 1.68B & 0.284  & 5.67 & 1.29 \\
30   & 512  & 1.68B & 0.290  & 5.65 & 1.47 \\
30   & 1024 & 1.68B & \textbf{0.291}  & \textbf{5.25} & 2.21 \\
\bottomrule
\end{tabular}
}}
\vspace{-3mm}
\end{table}

\noindent\textbf{Causal-Driven Stable Sampling.}
Based on the discussion in Sec.~\ref{sec_sampling_strategies}, we compare a simple top-k sampling strategy  with top-k=600. As shown in Fig.~\ref{fig_result_sample}, increasing of top-k to 4096 will enhance richness but also introduce structural instability.
Our Causal-Driven Stable Sampling mitigate this instability by injecting causal information during the progressive sampling while preserving detail richness. Additionally, incorporating confidence strategies within the sampler can further improve the results.

\section{Conclusion}
In this work, we explore the auto-regressive paradigm,~\ie, ``next-scale prediction'' for efficient text-to-image (T2I) synthesis. Our approach, STAR, predicts discrete feature maps in a scale-wise manner, guided by both global and local features extracted from a pretrained text encoder. Additionally, it employs normalized RoPE to prevent positional confusion across scales, thereby encoding token positions efficiently and enabling high-resolution training at a reduced computational cost. Furthermore, we analyze the sampling process within this scale-wise paradigm and propose a causal-driven sampling method to enhance image quality, achieving clearer structure and enriched detail. 

STAR achieves competitive performance in terms of fidelity and text-image alignment. Remarkably, it generates a high-quality 1024$\times$1024 resolution image in a highly efficient manner. STAR offers significant time advantages and produces detailed images compared to leading diffusion and previous AR models, presenting a promising new direction in the currently diffusion-dominated field of T2I generation.
{
    \small    \bibliographystyle{ieeenat_fullname}
\bibliography{main}
}

\clearpage
\setcounter{page}{1}
\maketitlesupplementary
\setcounter{section}{0}
\setcounter{figure}{0}
\setcounter{table}{0}

\renewcommand\thesection{\Alph{section}}

\section{Overview}
\label{sec:supp_overview}
This document provides supplementary materials for the
main paper. Specifically,~\cref{sec_supp_sample} presents more analysis of sampling strategy.
Sec.~\ref{sec_supp_attn} discusses the distribution of attention maps.
Supplemental visual results regarding STAR and other methods can be found at Sec.~\ref{sec_supp_visual}.

\section{Sampling Strategy}
\label{sec_supp_sample}
\subsection{Comparison with Current Strategies}
In the main text, we discuss how different sampling strategies impact results due to the inherent sampling stability issues in the scale-wise paradigm, which often involve a trade-off between image quality and diversity.

To further illustrate the importance of the proposed sampling method, we compare several strategies: a simple top-k=600 approach (the original setting in VAR~\cite{tian2024var}), a gumbel-noise-based method (Inject noise into the conditional probabilities of each scale through a gradually decreasing noise scheduler to generate randomness), and our learning-based strategy.

\begin{figure}[tbp]
\setlength{\abovecaptionskip}{0.1cm}
\setlength{\belowcaptionskip}{0.1cm}
  \centering
   \includegraphics[width=1.0\linewidth]{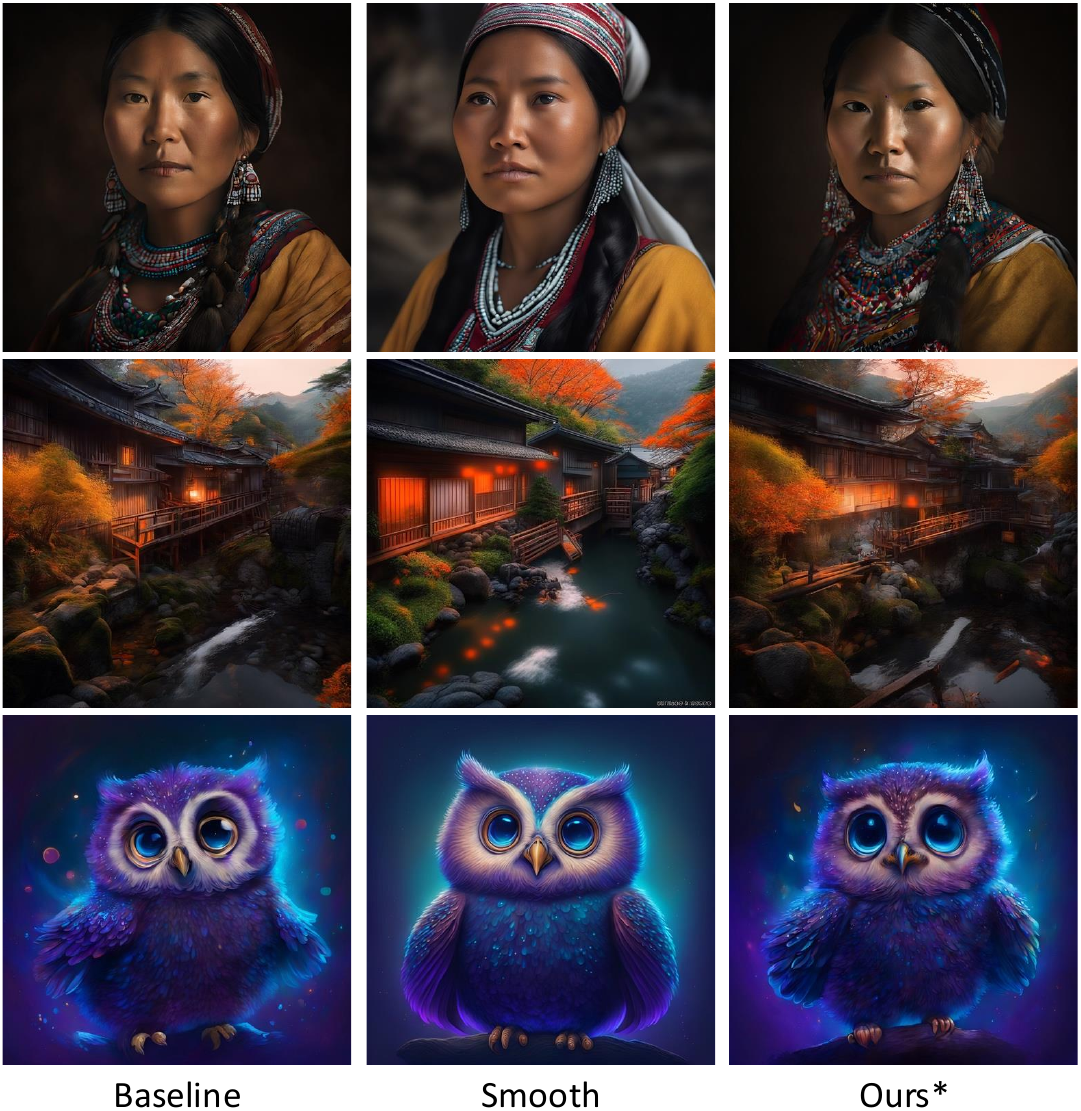}

   \caption{Images generated using different sampling strategies are presented: ``Baseline" employs a simple top-k=600 sampling, ``Smooth" uses gumbel-noise-based sampling, and ``Ours*" applies the proposed sampling method. The proposed method delivers richer image details and more stable image structures.}
   \label{fig:supp_sample}
\end{figure}

The experimental results demonstrate that the top-k=600 method leads to images lacking sufficient detail and produces unreliable results due to inconsistencies in token sampling directions. The gumbel-noise-based method tends to generate overly smooth images with missing details. As shown in~\cref{table_ab_sampler} and~\cref{fig:supp_sample}, the proposed sampling method achieves the best FID while maintaining strong text-image alignment.

\subsection{Analysis of Different Parts}
In the main text, we emphasize that a larger top-k is essential for enhancing image details, especially for 256-resolution images. However, for 1024-resolution images, increasing the top-k to 4096 introduces a degree of confusion (See~\cref{table_ab_sampler}). This may be attributed to the number of scales exceeding 256 for such images and the challenges in training and generating samplers at this resolution. By integrating the advanced sampling strategy discussed in the main text, this issue can be mitigated to some extent, while preserving detailed image features.

\begin{table}[h]
\caption{Comparison of different samplers on a 3k-image subset of MJHQ~\cite{li2024playground}. ``Ours" indicates the plain sampling strategy in~\cref{sec_sampling_strategies} of main text, ``Ours*" is the advanced approach mentioned in~\cref{sec_sampling_strategies_opt}, which showcases better balance between fidelity and text-image consistency.}
\label{table_ab_sampler}
\setlength{\tabcolsep}{2.5mm}{
\resizebox{1\columnwidth}{!}{
\begin{tabular}{ccccc}
\hline
Method   & top-k & CLIP-Score$\uparrow$ & FID$\downarrow$ & CMMD$\downarrow$ \\ \hline
Baseline & 600   & 0.289  & 22.60        & 0.368 \\
Smooth   & -     & 0.290  & 26.13        & 0.345 \\
Ours     & 600   & 0.290  & 22.08        & 0.347 \\
Ours     & 4096   & 0.289  & 23.75        & 0.396 \\
Ours*    & 4096  & 0.290  & 22.06        & 0.352 \\ \hline
\end{tabular}
}}
\end{table}
\begin{figure*}[t]
\begin{center}
\includegraphics[width=1\textwidth]{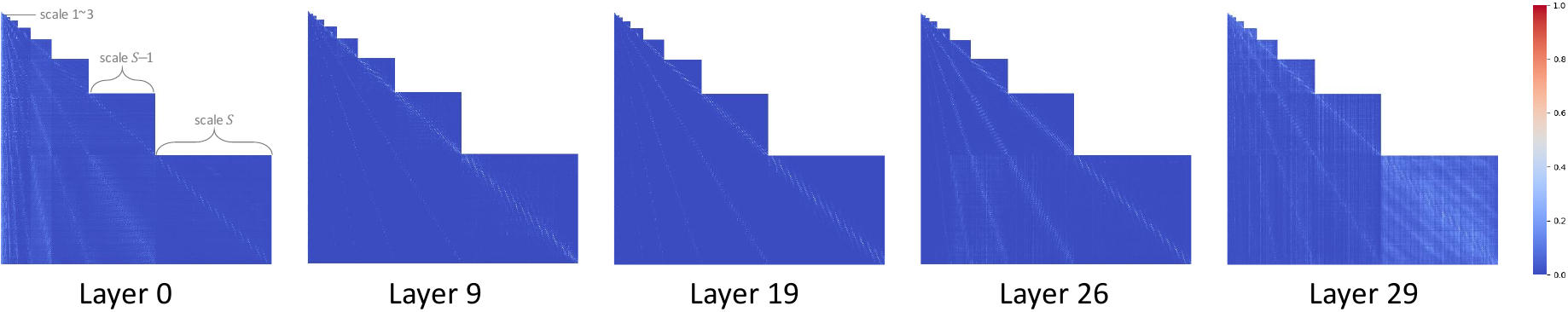}
\end{center}
\caption{Visualization of attention maps from different layers. It reveals tokens focus on local and aligned positions within scales, shifting to global attention in final layers, highlighting the need for fine-tuning in local window-based acceleration.}
\label{fig_supp_vis_attn_map}
\end{figure*}

\section{Additional Analysis on Attention Maps}
\label{sec_supp_attn}
During training, we dropped a portion of tokens from the last two scales to reduce training costs at 1024 resolution. Specifically, we adopted a local window-based strategy informed by observations of the attention map. As shown in~\cref{fig_supp_vis_attn_map}, we present an example of attention maps generated at 1024 resolution for each scale. The visualization reveals that for a specific token, its attention is predominantly focused on tokens within the same scale and tokens at relatively aligned positions in previous scales. This pattern becomes more pronounced in later scales. Such findings support the feasibility of using local window-based training acceleration methods. However, it is worth noting that in the final layers, attention tends to shift towards global information within the last scale, which could lead to potential performance degradation. Fine-tuning is required to fully recover the generative capabilities.

\section{Additional Visualize Results}
\label{sec_supp_visual}
\begin{figure*}
\begin{center}
\includegraphics[width=1\textwidth]{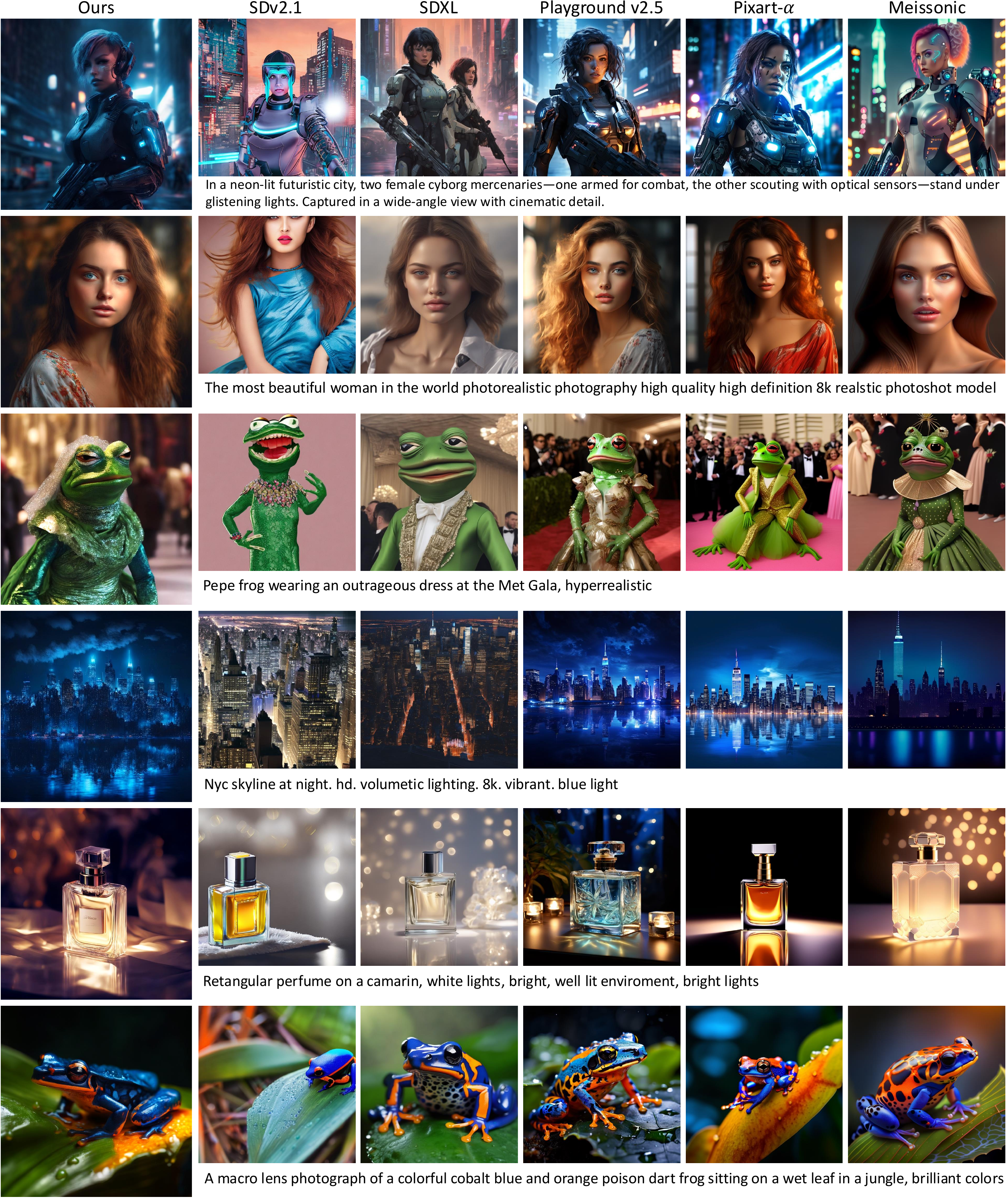}
\end{center}
\caption{Additional visual results of different methods.}
\label{fig_supp_vis1}
\end{figure*}

\begin{figure*}
\begin{center}
\includegraphics[width=1\textwidth]{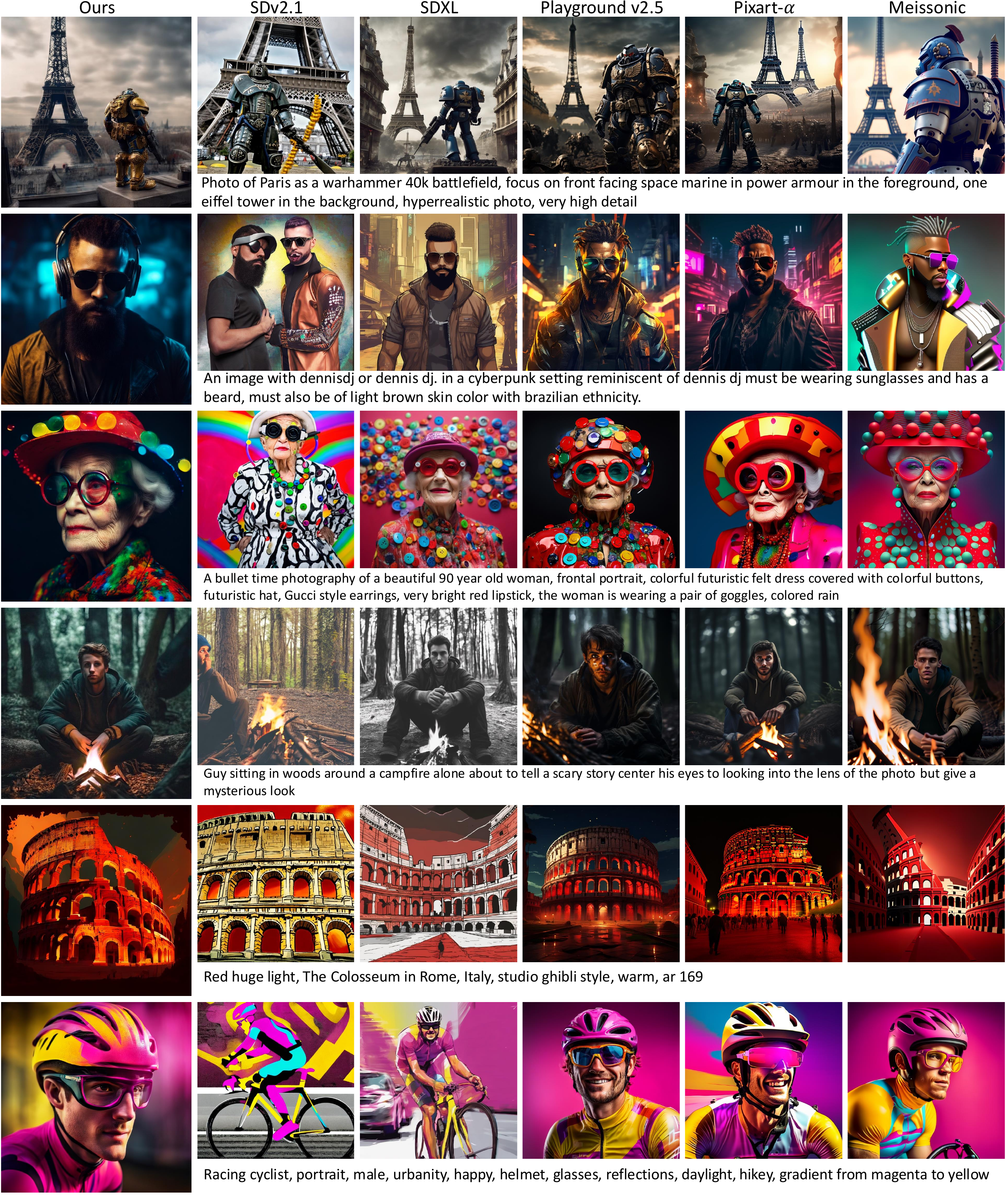}
\end{center}
\caption{Additional visual results of different methods.}
\label{fig_supp_vis2}
\end{figure*}

\begin{figure*}
\begin{center}
\includegraphics[width=1\textwidth]{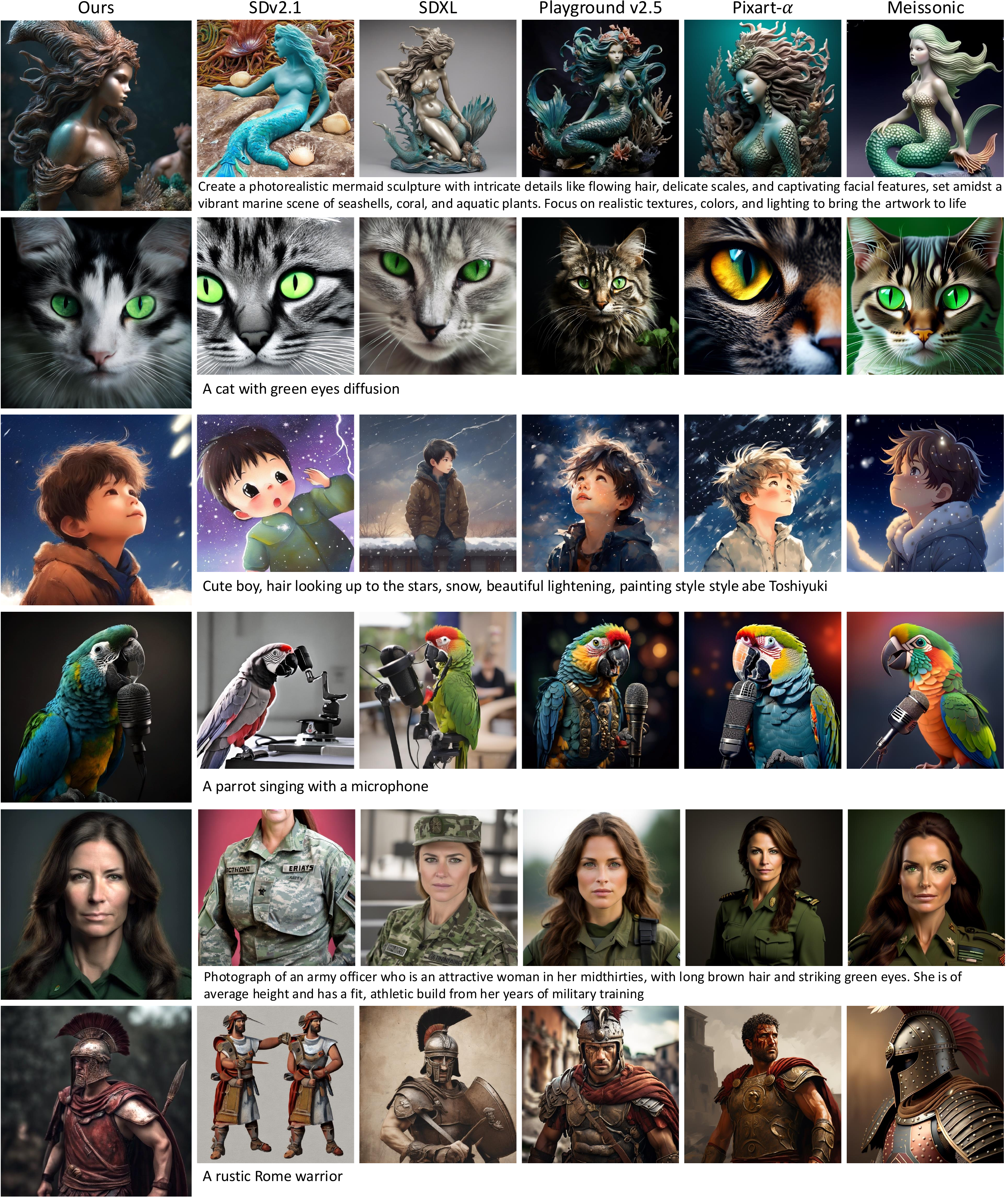}
\end{center}
\caption{Additional visual results of different methods.}
\label{fig_supp_vis3}
\end{figure*}

In the main text, we presented visual comparisons with SOTA methods in~\cref{fig:comp_sota}. Here, we provide additional visual results. As shown in~\cref{fig_supp_vis1} to~\cref{fig_supp_vis3}, STAR can generate images with diverse types and styles while achieving significantly higher efficiency compared to current diffusion models.

\end{document}